\documentclass[times, review, 10pt]{elsarticle}

\usepackage{amssymb}
\usepackage{color}
\usepackage{multirow}
\usepackage{amsmath}
\usepackage{amssymb}
\usepackage{amsfonts}
\usepackage{bm}
\usepackage{amssymb}
\usepackage{pifont}

\usepackage{graphicx}

\usepackage{booktabs}
\usepackage{multirow}
\usepackage{array}

\usepackage[colorlinks,linkcolor=blue]{hyperref}

\usepackage{url}

\usepackage[caption=false,font=normalsize,labelfont=sf,textfont=sf]{subfig}

\usepackage{textcomp}
\usepackage{verbatim}

\usepackage{amsmath}

\journal{Pattern Recognition}

\begin{document}

\begin{frontmatter}



\title{TACoS: Weakly Supervised Learning of Two-Dimensional Materials from Scribble Annotations to Precise Segmentation}

\author[1,3]{Jiabei Chen}

\author[1,3]{Liping Zhang}

\author[2,3]{Jiang-Bin Wu}

\author[2,3]{Zhongming Wei}

\author[1]{Enhao Ning}

\author[1,3]{Su Yan}

\author[1,3]{Weijun Li}

\author[2,3]{Ping-Heng Tan}

\author[1,3]{Xin Ning\corref{cor1}}
\ead{ningxin@semi.ac.cn}

\affiliation[1]{organization={AnnLab, Institute of Semiconductors, Chinese Academy of Sciences},
                city={Beijing},
                postcode={100083}, 
                country={China}}

\affiliation[2]{organization={State Key Laboratory of Semiconductor Physics and Chip Technologies, Institute of Semiconductors, Chinese Academy of Sciences},
                city={Beijing},
                postcode={100083}, 
                country={China}}

\affiliation[3]{organization={Center of Materials Science and Optoelectronics Engineering \& School of Integrated Circuits, University of Chinese Academy of Sciences},
                city={Beijing},
                postcode={100049}, 
                country={China}}

\cortext[cor1]{Corresponding author}

\begin{abstract}
The precise pixel-level localization of two-dimensional material flakes is crucial for high-throughput screening. However, traditional fully supervised methods rely on dense annotations, which are costly and time-consuming, severely limiting the practical deployment of segmentation models. This paper proposes Tree-based Asymmetric Contrast Segmentation (TACoS), a specialized scribble segmentation framework tailored for two-dimensional materials. First, we design a unified framework that integrates semi-supervised consistency learning with structured tree energy constraints. This framework comprises two core components: an unlabeled weak-strong distribution alignment module and a tree energy regularization module. The former employs cosine consistency constraints to enhance prediction alignment across views. Meanwhile, the latter utilizes minimum spanning trees to establish pixel affinity relationships and generate structure-aware soft pseudo labels for online semantic guidance. Next, we introduce asymmetric regional contrast learning. This approach fuses high-confidence predictions from the weak augmentation branch with scribbles to form augmented labels, and construct category prototypes in the representation space. Simultaneously, we prioritize contrastive constraints on challenging pixels in boundary-unlabeled regions. This strategy enhances intra-class cohesion and inter-class separation at the representation level, effectively reducing category confusion in low-contrast edges and complex backgrounds. Experiments conducted on the constructed graphene and MoS$_2$ datasets demonstrate that our method TACoS achieves over 96\% of fully supervised performance using less than 0.6\% annotated data. Furthermore, it exhibits superior structural coherence and boundary stability in scenarios with weakly contrasting edges and complex backgrounds, providing an efficient and scalable solution for automated high-throughput screening of two-dimensional material flakes.
\end{abstract}



\end{frontmatter}

\section{Introduction}\label{Sec:introduction}

Since graphene was successfully isolated \cite{Geim2007}, two-dimensional materials have rapidly become a major research focus in condensed-matter and materials physics. Experimentally, mechanical exfoliation \cite{Novoselov2004} is widely used to prepare two-dimensional material samples due to its simplicity and high crystalline quality. Taking advantage of thin-film interference \cite{Blake2007}, exfoliated flakes exhibit obvious reflectance contrast and clear morphological cues under visible light, which provides a physically observable basis for pixel-level flake segmentation from microscopic images.

In recent years, semantic segmentation has been successfully applied to automated identification of two-dimensional material flakes \cite{Dong2022,Sterbentz2021}. When training images are sufficiently annotated, models can learn optical contrast, edge gradients, and texture patterns of target regions, enabling precise segmentation at pixel-level. However, in real laboratory settings, images are often large, have complex backgrounds, and contain diverse impurities. Pixel-level annotation typically requires domain experts investing substantial time and effort. Consequently, the substantially increasing annotation and maintenance cost directly limits training efficiency and scalability. Purely dense-annotation-based fully supervised solutions are therefore difficult to deploy at scale in routine experimental workflows.

\begin{figure*}[!htbp]
  \centering
  \includegraphics[width=\linewidth]{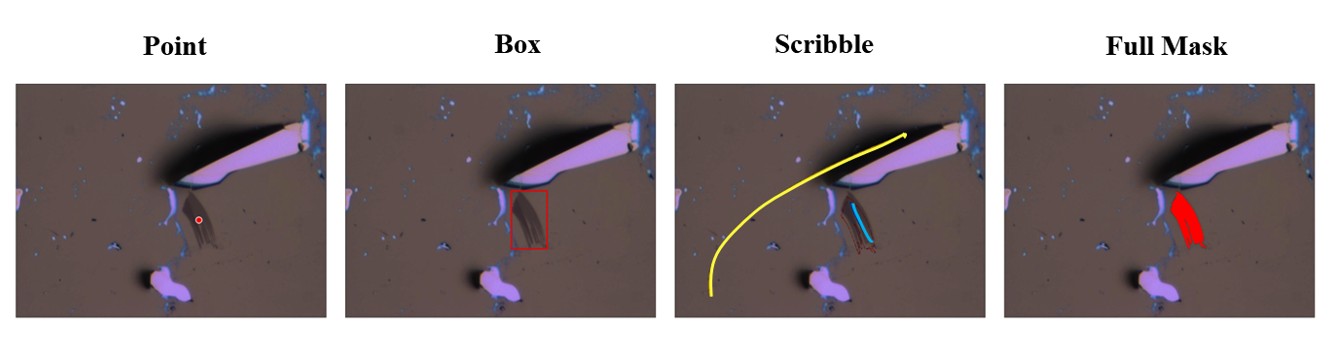}
  \caption{\textbf{Annotation types illustration.} We visualize four common annotation regimes for flake segmentation: point, box, scribble, and full mask. Point and box annotations are placed on the foreground flake region, while scribble provides sparse strokes along the object structure; the full mask serves as the dense supervision reference.}
  \label{fig:anno_types}
\end{figure*}

Weakly supervised learning offers an alternative to reduce annotation cost. As shown in \autoref{fig:anno_types}, scribble lies between point and full mask, allowing annotators to draw coarse outlines rather than precise contours. With consistency learning or perturbation strategies, models can exploit information from unlabeled regions via sparse scribbles to recover full segmentation capability. Although scribble cost slightly more than points and have only limited information gain compared to weaker supervision forms like box, they naturally encode shape priors and often achieve a favorable trade-off between performance and annotation efficiency. It should be noted that existing scribble-supervised segmentation studies mainly focus on natural scenes \cite{Lin2016} and medical images \cite{Can2018}, with no systematic studies reported on two-dimensional material microscopic data.

Existing scribble-supervised segmentation methods can be roughly categorized into regularization learning \cite{Pan2021,Pan2024}, pseudo-label learning \cite{Liang2022,Wu2023,Wu2025,Zhang2022AAGNN}, and consistency learning \cite{Su2023}. However, these approaches all exhibit shortcomings in the context of two-dimensional thin-section material microscopic images. Regularization learning propagates supervision signals by constraining label consistency between neighboring pixels. However, in two-dimensional material segmentation scenarios, its emphasis on low-level material features — such as color and grayscale — is susceptible to background texture interference, making it difficult to effectively integrate with high-level semantic boundary judgments. Pseudo-label learning attempts to densify supervision via self-training, yet when flake and background colors are similar and the thickness or interference color varies, the pseudo-labels generated by the model near semantic boundaries can be systematically biased and errors may accumulate over iterations. Consistency learning aligns predictions under different perturbations to regularize the model, but it mainly enforces output stability rather than directly improving boundary discriminability. At the core lies the spatial nature of scribbles: annotators typically mark high-confidence interior regions, leaving boundary pixels almost unsupervised. Boundaries are precisely the hardest part to learn because of geometric sparsity (minuscule proportion relative to interior areas), feature ambiguity (resemblance to substrate textures, contaminants, or crease shadows), and imaging degradation (uneven lighting, low-contrast transition zones). Constrained by the aforementioned bottlenecks, existing methods often compromise between boundary accuracy and regional consistency, making them ill-suited for the task of segmenting two-dimensional material sheets.

Therefore, this paper proposes TACoS (Tree-based Asymmetric Contrast Segmentation), the first annotation segmentation framework specifically designed for two-dimensional materials. It unifies weak-strong consistency learning, structured regularization, and online pseudo-label semantic guidance within a single end-to-end training framework. This approach enables efficient learning in unlabeled regions at low annotation cost while notably enhancing boundary and structural stability. TACoS consists of three tightly coupled key designs: Unlabeled Weak–Strong Distribution alignment (UWSD), Tree Energy Regularization (TER), and Asymmetric Regional Contrastive Learning (ARCL). Specifically, UWSD enforces consistency constraints on prediction distributions between weak/strong augmentation views only on unlabeled pixels, providing dense training signals for the dominant unlabeled regions. TER builds a minimum spanning tree on backbone features, and uses tree filtering to generate a structure-aware soft reference, imposing structured regularization on unlabeled pixels to suppress foreground fragmentation and connected-component drift. ARCL completes unlabeled pseudo labels using high-confidence predictions from the weak augmentation branch, combines them with scribble annotations to form augmented labels, and applies contrastive constraints on strong-augmentation-branch representations. By turning boundary learning into a contrastive learning problem, ARCL enhances boundary-discriminative representations without additional boundary labels, compensating for the lack of direct class-level guidance at boundaries under sparse supervision.

Experimental results indicate that even with less than 0.6\% annotated data, this method achieves over 96\% of full-supervision performance while exhibiting superior regional coherence and boundary stability in visualization results.

In summary, the main contributions of this paper are as follows:
\begin{itemize}
\item For the first time, we introduce sparse supervision into the segmentation of two-dimensional materials, proposing a unified framework called TACoS that integrates semi-supervised consistency learning with structured tree energy constraints. This addresses the lack of semantic guidance and boundary degradation in unlabeled regions of thin two-dimensional material samples under sparse supervision.
\item Design an asymmetric region contrast learning module to construct class prototypes in the representation space and perform region-level contrast learning. Simultaneously, focus contrast constraints on challenging pixels in unlabeled boundary regions to enhance intra-class cohesion and inter-class separation at the representation level, thereby reducing class confusion in scenarios with weak contrast edges and complex backgrounds.
\item Through quantitative metrics and visual analysis, TACoS has been thoroughly validated to demonstrate superior structural consistency and boundary stability in scenarios involving weak contrast edges and complex backgrounds.
\end{itemize}

The remainder of this paper is organized as follows. \autoref{Sec:related_work} reviews related work on automatic segmentation of two-dimensional material micrographs, semi-supervised segmentation, and contrastive learning segmentation, outlining the relationship between our method and existing approaches along with improvements. \autoref{Sec:method} details the overall framework, training objectives, and implementation specifics of TACoS under a unified network setting. \autoref{Sec:experiments} reports experimental results and analysis. To ensure representativeness and reproducibility of comparisons, we adopt DINOv2 \cite{Oquab2024} as the unified backbone network, DPT \cite{Ranftl2021} as the segmentation head, and align training budgets. We select typical scribble-supervised methods with different mechanisms as comparison baselines, including A$^2$GNN \cite{Zhang2022AAGNN}, URSS \cite{Pan2021}, CC4S \cite{Pan2024}, AGMM \cite{Wu2023}, and SASFormer \cite{Su2023}. We also include FlashInternImage \cite{Xiong2024} with a DCNv4 backbone as a strong reference point for comparative training, and conduct ablation studies to evaluate the individual contributions and coupling gains of UWSD, TER, and ARCL. Finally, \autoref{Sec:conclusion} concludes the paper and discusses future directions.

\section{Related Work}\label{Sec:related_work}

\subsection{Optical microscopic Images of Two-Dimensional Materials and Automated Segmentation}

Deep learning has been widely used for automated segmentation of two-dimensional material flakes, achieving progress in instance segmentation \cite{Dong2022,Masubuchi2020} and semantic segmentation \cite{Uslu2024,Yang2024,Zhang2022Coatings}. For instance segmentation, Masubuchi \cite{Masubuchi2020} built a dataset of over 2,100 optical microscopic images covering BN, graphene, MoS$_2$, and WTe$_2$, and integrated Mask R-CNN \cite{He2017} into an electric stage microscope system to enable an end-to-end workflow for automated scanning and flake instance segmentation. For semantic segmentation, Zhang \cite{Zhang2022Coatings} systematically benchmarked multiple semantic segmentation networks on public data and proposed 2DU2-Net, providing a representative baseline for model selection and a evaluation insight for low-contrast and fragmented morphologies. Overall, public datasets still suffer from limited scale and diversity, while pixel-wise mask annotation is expensive and expert-dependent, limiting generalization and deployment. 

To mitigate this, Yan \cite{Yan2025} constructed a large-scale dataset with 7,454 images and about 30,000 annotated flake regions across multiple substrates and imaging conditions, and proposed a batch-wise self-evolving annotation ecosystem that integrates active learning, semi-supervised learning, and foundation models to reduce manual cost, establishing a stronger basis for generalizable training and evaluation.

\subsection{Scribble-Supervised Semantic Segmentation}

Scribble-supervised semantic segmentation aims to express the shape features of target regions with low annotation cost, thereby accomplishing segmentation tasks more efficiently. Since ScribbleSup \cite{Lin2016} first extended point-level annotations to scribbles, related methods have continuously evolved. It is worth noting that many representative works were proposed under broader sparse-annotation settings (e.g., point annotations, bounding boxes, or mixed weak annotations), but their core mechanisms are also transferable to scribble supervision. 

A$^2$GNN \cite{Zhang2022AAGNN} propagates pseudo labels on a relational graph and suppresses weak-supervision bias using soft-edge weighting and attention propagation. Although this method can leverage edge priors to suppress errors, the effectiveness of its soft edge weights drops sharply under low contrast conditions, making it difficult for pseudo-labels to precisely align with contours. URSS \cite{Pan2021} combines uncertainty constraints with random-walk diffusion, excluding pseudo boundaries via entropy minimization to enhance region consistency. CC4S \cite{Pan2024} further improves pseudo label quality by introducing color regularization on this foundation. While these methods perform well in natural scenes, in two-dimensional material images where the optical features of the foreground and background are similar, random walks are highly prone to crossing semantic boundaries, resulting in pseudo-label bias. TEL \cite{Liang2022} uses a minimum spanning tree to model low-level color and high-level semantic similarity. However, the extremely low discriminative power of low-level features interferes with the topological construction of its tree structure, compromising edge integrity. AGMM \cite{Wu2023} and AGMM++ \cite{Wu2025} employ class-conditional Gaussian mixture models and online expectation-maximization processes to precisely model and refine feature distributions. However, due to partial overlap between foreground and background features in the feature space, they struggle to accurately separate features on either side of the boundary. SASFormer \cite{Su2023} utilizes hierarchical attention to propagate labeled region information and employs an affinity consistency loss to constrain predictions. However, in the absence of considerable visual cues, attention tends to diverge, and boundary pixels lack direct supervision, leading to the accumulation of mislabeling bias. Although the aforementioned methods have achieved considerable results in propagation denoising and online soft supervision generation, under low contrast and complex substrates in two-dimensional-material microscopic images they can still suffer from accumulating pseudo-label bias, and boundary pixels lack direct supervision, making contour quality hard to improve reliably.

In contrast, the TACoS method proposed in this paper provides dense unlabeled training signals through weak-strong alignment, introduces a regularization strategy that includes tree-structure constraints, enhances structural stability, and prevents the propagation of errors from false labels. Furthermore, a region-level contrastive learning module is introduced to increase the distance between the specimen and the substrate background as well as other impurities in the feature space. This effectively addresses the issue of indistinguishable boundaries caused by low contrast, thereby improving the quality of the specimen outline without increasing annotation costs.

\subsection{Semantic Segmentation with Semi-Supervised and Contrastive Learning}

Semi-supervised semantic segmentation trains with a small amount of labeled data alongside a large volume of unlabeled data. The key is to construct reliable supervision for unlabeled pixels and prevent bias amplification. Classic approaches like CCT \cite{Ouali2020} enhance training stability by constraining unlabeled regions through cross-consistency constraints across multi-branch and multi-perturbation predictions. CPS \cite{Chen2021} reinforces consistency and extends effective training signals by employing dual networks mutually supervised with pseudo labels. In addition, the recently proposed B$^3$CT \cite{liang2026} further designs a three-branch learning architecture that achieves domain-robust semantic segmentation by mining unlabeled target signals, providing a new paradigm for leveraging unlabeled information under complex feature distributions. Meanwhile, contrastive learning enhances intra-class compactness and inter-class separation in the representation space. For instance, CIPC \cite{Wang2021} introduces global context through cross-image pixel-level comparisons, while STEGO \cite{Hamilton2022} distills semantic clusters under fully unlabeled conditions by leveraging pre-trained self-supervised features to capture consistent structural relationships within and across images. Combined with contrastive loss and structured graph optimization, it achieves consistent and detailed unsupervised semantic segmentation results.

Although this paper focuses on scribble-based weak supervision, the utilization of a large number of unlabeled pixels aligns closely with the semi-supervised setting. we propose the single-stage framework TACoS, comprising three modules, each introducing its reference source and design rationale: UWSD is inspired by the weak augmentation pseudo-label and strong-augmentation consistency alignment of the classic semi-supervised model FixMatch \cite{Sohn2020} and leverages data-augmentation strategies from UniMatch V2 \cite{Yang2025} to better utilize the large proportion of unlabeled pixels. TER is motivated by the minimum spanning tree modeling in TEL \cite{Liang2022}, using tree structures to characterize pixel affinity relationships and a structure-aware soft reference to suppress fragmentation and drift. Inspired by contrastive learning model, ARCL draws from the region-prototype and hard-sampling mechanism of ReCo \cite{Liu2022} and the boundary pixel pair contrast with confidence-guided prediction in GCBL \cite{Qiu2024}: it completes supervision with high-confidence predictions from the weak augmentation branch and drives regional contrastive learning, focusing representation constraints on uncertain and boundary neighborhoods to improve boundary discriminability.

\section{Method}\label{Sec:method}

\subsection{Task Setup}
This paper targets semantic segmentation of two-dimensional material flakes in optical microscopic images, adopting a default binary classification setup, the foreground representing flake regions, while the background encompasses blank substrates, multilayer non-target areas, impurities, and other elements. Let the input image be

\begin{equation}
x\in\operatorname{R}^{H\times W\times C}.
\end{equation}

Segment network output pixel-level category probability maps

\begin{equation}
p\in[0,1]^{H\times W\times K},\ \ K=2.
\end{equation}

In our dataset, dense masks take values in {0, 1}; for the scribble setting, the label mask uses {0, 1, 255}, where 255 denotes unknown pixels. We partition the pixel set $\Omega$ into a labeled pixel set $\Omega_L$ and an unlabeled pixel set $\Omega_U$, such that $\Omega_L \cup \Omega_U = \Omega$ and $\Omega_L \cap \Omega_U = \emptyset$. This explicit modeling of pixel-level labeling deficiencies serves as the core starting point for the subsequent design of TACoS.

\subsection{TACoS: Unified Scribble Segmentation Framework}

\begin{figure*}[!htbp]
    \centering
    \includegraphics[width=0.8\textwidth]{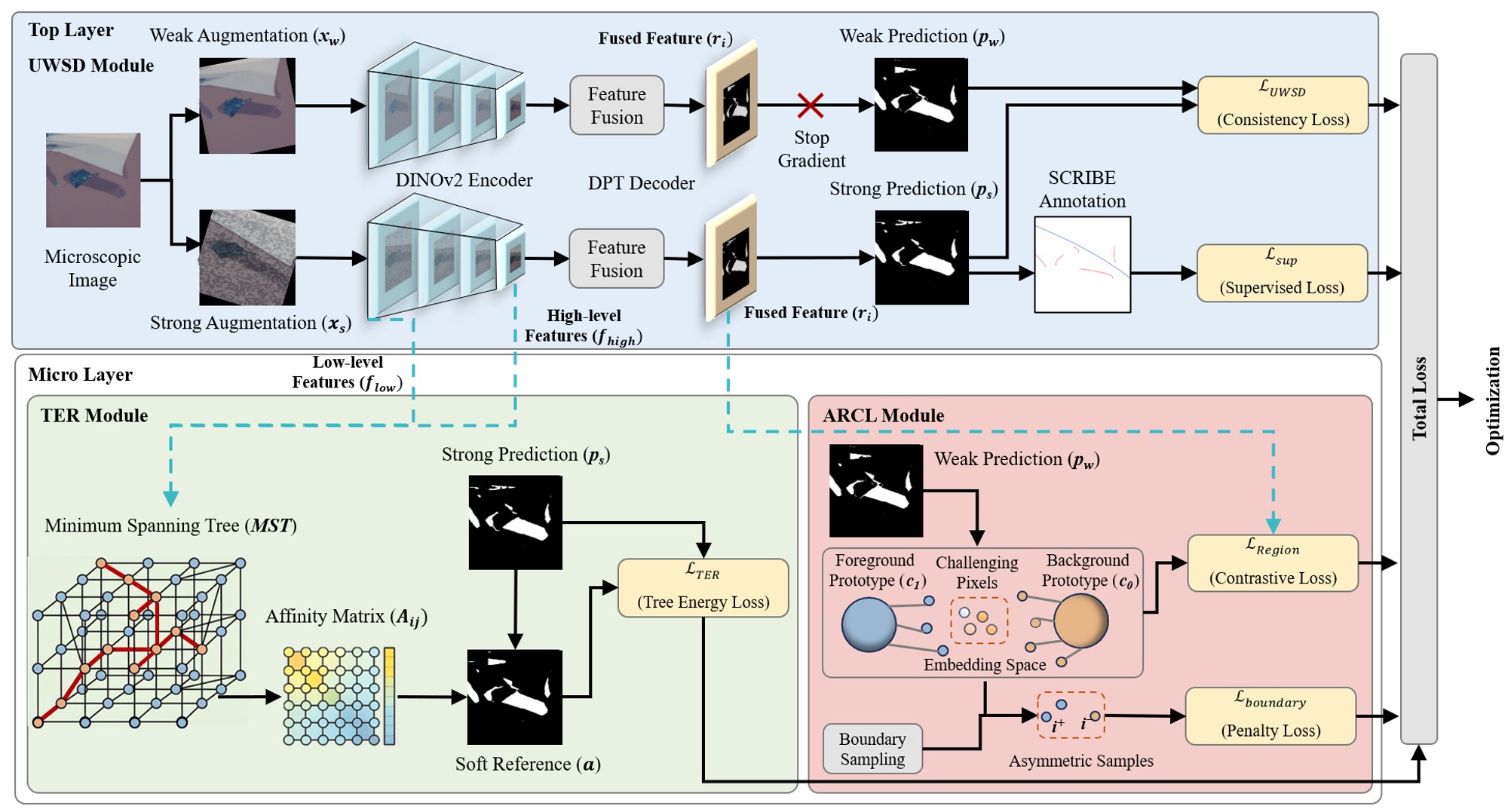}
    \caption{\textbf{Overview of the proposed TACoS framework.} The top layer performs weak–strong consistency learning with a shared DINOv2–DPT segmentation network, where weak predictions provide stop-gradient pseudo supervision while scribbles supply supervised signals. The bottom layer introduces two complementary structural modules. TER models pixel relations via a minimum spanning tree in the encoder feature space and generates a structure-aware soft reference for tree energy regularization. ARCL enhances representation separability by performing prototype-based region contrastive learning on challenging pixels and applying asymmetric boundary sampling to enforce boundary-aware penalties. All modules are jointly optimized under a unified objective.}
    \label{fig:tacos_framework}
\end{figure*}

As shown in \autoref{fig:tacos_framework}, the proposed TACoS framework adopts a dual-branch joint optimization paradigm to recover complete segmentation masks for two-dimensional materials from extremely sparse scribble annotations. Given an input image, the network feeds its weak and strong augmentation views into an encoder-decoder with shared weights. To fully leverage information from unlabeled regions, TACoS tightly couples three core modules during end-to-end training: The Unlabeled Weak-Strong Distribution Alignment (UWSD) module forces the strong enhancement branch's predictions to align outputs from the weak branch, providing dense pixel-level consistency supervision signals for unlabeled areas. The Tree Energy Regularization (TER) module combines low-level details with high-level semantic features to construct minimum spanning trees and perform two-stage tree filtering, generating soft references with global awareness that impose strict spatial structural regularization constraints on unlabeled pixels. The Asymmetric Region Contrast Learning (ARCL) module fuses online pseudo labels with sparse scribbles and enforces region-level asymmetric contrast constraints in the feature space of the strong enhancement branch, substantially improving boundary discrimination accuracy. Ultimately, these three constraints are jointly optimized with standard scribble loss within a unified objective function, achieving a stable transition from sparse supervision to high-quality dense prediction. Unlike multi-stage approaches, TACoS employs a strict single-stage design. This means that all optimization objectives are jointly minimized in a single continuous training process, completely eliminating the need for offline pseudo-label generation, iterative model retraining, and separate fine-tuning stages.

\subsubsection{UWSD: Unlabeled Weak–Strong Distribution Alignment}

In weakly supervised semantic segmentation, the key to effectively utilizing unlabeled pixels lies in constructing a robust consistency regularization framework. Drawing inspiration from UniMatch V2's design, this paper adopts a weak-strong dual-branch architecture: leveraging the model's high-quality predictions on images with simple transformations (weak augmentations) to guide its learning of representations under complex transformations (strong augmentations) of the same images. This training approach not only substantially enhances the model's robustness against noise and perturbations but also challenges it to discover invariant feature representations across different image perspectives.

To extend consistency regularization from the image level to the pixel level, given a microscopic image $x$, we construct the weak augmentation view and the strong augmentation view within the same cropped field of view during training:

\begin{equation}
x^w=T_{geo}(x),\ \ x^s=T_{app}(T_{geo}(x)),
\end{equation}
here, \(T_{geo}(\cdot)\) denotes geometric transformations, including random scaling, random cropping to fixed dimensions, and random horizontal flipping. Input \(x\) undergoes this transformation to generate the weak augmentation branch input \(x^w\), with the scribble annotation label \(y\) transformed synchronously. Subsequently, the strong augmentation branch further applies appearance perturbations \(T_{app}(\cdot)\) relevant to microscopic imaging on top of this geometric transformation: it performs color jittering (brightness / contrast / saturation / hue) with a certain probability, and applies grayscaling and random blurring with a low probability. This simulates imaging variations such as fluctuations in light intensity, color temperature, and slight defocusing. Thus, the input $x$ undergoes both geometric transformation and appearance perturbation to generate the input $x^s$ for the strong augmentation branch. The geometrically transformed label $y$ serves as the direct supervision signal for this branch, while $x^w$ and $x^s$ are utilized for consistency regularization learning.

To ensure consistency regularization acts solely on the unlabeled pixel set $\Omega_U$. In unlabeled regions, UWSD employs the prediction distribution of the weak augmentation view as a soft objective, constraining the predictions of the strong augmentation view to align with it. This suppresses the propagation of apparent perturbation bias from strong augmentation onto unlabeled pixels.

Let \(d(p,q) = 1 - \cos(p,q)\), then we have

\begin{equation}
d(p,q) = 1-\frac{\langle p,q\rangle}{\|p\|_2\,\|q\|_2+\epsilon}.
\end{equation}

We use the weak-augmentation prediction as the alignment target and stop its gradients from backpropagating.

\begin{equation}
\mathcal{L}_{\text{UWSD}}=\frac{1}{|\Omega_U|}\sum_{i\in\Omega_U}d\!\left(\mathrm{sg}\!\left(p_w(i)\right),\,p_s(i)\right).
\end{equation}

Stopping gradient backpropagation is necessary because the weak-augmentation branch only provides a relatively stable soft target and is not updated by this loss. This prevents the weak and strong augmentation branches from chasing each other, which could otherwise cause distribution drift or training collapse. This is particularly important under sparse supervision, since $\Omega_U$ typically accounts for the vast majority of pixels in an image.

The weak/strong augmentation views share the same set of network parameters and utilize a common encoder and decoder, with their branching structure following the overall paradigm of UniMatch V2.

\subsubsection{TER: Tree-Energy Regularization}

UWSD can provide dense consistency signals on $\Omega_U$, but it remains fundamentally an alignment at the distribution level without explicitly constraining structural relationships between pixels. To address this, we introduce TER, which constructs a minimum spanning tree (MST) in feature space to capture pixel affinity relationships and generates structure-aware online soft references based on this. To address the optical properties of two-dimensional material flakes, we extract shallow and deep feature representations from DINOv2 to construct a dual-tree architecture. The shallow features capture microscopic textures and color gradients to anchor boundaries, while the deep features encode high-level semantics to suppress background noise. This dual-tree mechanism ensures that the generated soft references possess both microscopic geometric accuracy and macroscopic semantic consistency.

Let the low-level (Layer 2) and high-level (Layer 11) features extracted from the strong-augmentation branch be

\begin{equation}
f_{low}\in\operatorname{R}^{H_l\times W_l\times D_l},\ \ f_{high}\in\operatorname{R}^{H_h\times W_h\times D_h}.
\end{equation}

Based on the aligned features, we define a pixel graph where all pixels form the vertex set and the feature distances between adjacent pixels form the edge set. By successively removing edges with the largest weights while preserving the graph's connectivity, we can construct a minimum spanning tree:

\begin{equation}
\mathcal{G}=\operatorname{MST}\left(\Omega,\mathrm w_{ij}=\parallel f(i)-f(j)\parallel_2\right),
\end{equation}
where $f(\cdot)$ takes the values $f_{low}$ and $f_{high}$, respectively, we construct two minimum spanning trees $\mathcal{G}_{low}$ and $\mathcal{G}_{high}$ in the low-level and high-level feature spaces, respectively.

To capture long-range dependencies between pixels, we map tree distances to an affinity matrix:

\begin{equation}
A_{ij}=\exp\left(-\frac{D_{ij}}{\sigma_{aff}}\right),
\end{equation}
where $D_{ij}$ denotes the sum of path edge set distances between any two points in the minimum spanning tree $\mathcal{G}$, and $\sigma_{aff}$ represents the scale parameter.

Given a strong augmentation branch prediction $p_s$, we apply the tree filter operator $\mathcal{T}(\cdot)$ to the affinity matrix $A$, yielding a structure-aware soft reference:

\begin{equation}
a=\mathcal{T}_{high}(\mathcal{T}_{low}\left(\Omega,\ p_s,A)\right),\ \ a(i)\in[0,1]^K,
\end{equation}
among these, the filtering operator $\mathcal{T}$ represents the result of normalizing the weighted sum of the affinity $A_{ij}$ across all image pixels $\Omega$, weighted by the prediction probability $p_s$. This process can be viewed as adaptive smoothing of the prediction distribution under pixel affinity constraints, ensuring that highly similar pixels remain consistent in the prediction space. Unlike offline-generated pseudo labels, $a$ is computed online from the current feature structure and current prediction, dynamically updated at each iteration. This provides structure-aware semantic guidance for $\Omega_U$ without increasing manual annotation costs, while effectively suppressing connected domain drift and prediction bias.

On the unlabeled pixel set $\Omega_U$, we minimize the discrepancy between the strong-augmentation-branch prediction and the soft reference:

\begin{equation}
\mathcal{L}_{TER}=\frac{1}{\mid\Omega_U\mid}\sum_{i\in\Omega_U}\parallel p_s(i)-\operatorname{sg}(a(i))\parallel_1.
\end{equation}

\subsubsection{ARCL: Asymmetric Regional Contrast Learning}

Although UWSD and TER provide dense consistency signals and structured regularization on the unlabeled pixel set $\Omega_U$, the representational separability at the category boundary may still be insufficient for models when dealing with weakly contrasting edges and complex backgrounds. To this end, we introduce ARCL, imposing constraints on boundary neighborhoods from both the representation space and decision space. On one hand, performing region-level contrastive learning on deep decoder features. On the other, applying an exclusion penalty based on asymmetric distance sampling to boundary neighborhoods on strong augmentation branch logits. Both share the same filtering logic for valid pixels and difficult samples, thereby forming a consistent boundary enhancement mechanism.

We first construct the augmented label \(\widetilde{y}\) using the weak augmentation branch prediction \(p_w\). For pixel \(i \in \Omega\):

\begin{equation}
\widetilde{y}(i)=\left\{\begin{matrix}y(i),&i\in\Omega_L,\\\arg{\max}_kp_w^k(i),&i\in\Omega_U\land{\max}_kp_w^k(i)\geq\tau_w,\\\mathrm{255},&\mathrm{otherwise},\\\end{matrix}\right.,
\end{equation}
here, $\tau_w$ corresponds to the weak confidence threshold. In this binary classification segmentation task, $k$ takes the values 0 or 1. This yields the effective pixel mask.

\begin{equation}
m_V\left(i\right)=\mathbb{I}\left[\widetilde{y}\left(i\right)\neq\mathrm{255} \right].
\end{equation}

Unlike label generation, sampling and weight computation in contrastive learning are entirely based on strong augmentation branch predictions. A stricter strong confidence threshold $\tau_s$ is employed to filter out high-confidence samples. We focus the candidate set on the collection $\mathcal{S}$ of challenging pixels that are classified as foreground by augmented labels but exhibit sharp confidence drops or misclassification under strong perturbations.

\begin{equation}
\mathcal{S}=\left\{ i \;\middle|\; m_V(i)=1 \;\land\; \max\limits_{k} p_s^{k}(i) < \tau_s \right\}.
\end{equation}

For each pixel $i \in \Omega$, let $\mathbf{r}_i$ denote the depth-aggregated feature extracted by the segmentation network from the strong augmentation view. On $\mathcal{S}$, the contrastive loss is defined as:

\begin{equation}
\mathcal{L}_{\text{Region}}
=-\frac{1}{|\mathcal{S}|}\sum_{i\in\mathcal{S}}\log\frac{\exp\!\left(\langle r_i, c_{\tilde{y}(i)}\rangle/T\right)}{\sum_{k}\exp\!\left(\langle r_i,c_k\rangle/T\right)},
\end{equation}
here, $c_k$ denotes the mean feature vector of valid pixels in class $k$ within the current input batch (the class $k$ regional prototype), while $c_{\tilde{y}(i)}$ represents the target class prototype corresponding to pixel $i$ based on the augmented label $y(i)$, and $T$ represents the temperature coefficient. This loss function promotes intra-class cohesion and expands inter-class separation at the representation level.

Meanwhile, within the decision space, ARCL directly imposes constraints on the boundary neighborhoods of strong branch logits $z_s$. Since features in boundary neighborhoods are typically highly smooth, traditional symmetric local sampling tends to lead to conflicting update signals or misdirection. To address this, we adopt an asymmetric distance-based boundary search strategy.

The label map for boundary supervision is denoted as $y_B$. In the implementation, a full mask is uniformly applied to extract boundaries. The morphological gradient based on the label map $y_B$ yields the geometric boundary set $\mathcal{B}$. In practice, we select boundary pixels for supervision only within the set of challenging pixels $\mathcal{S}$, i.e., $\mathcal{E} = \mathcal{B} \cap \mathcal{S}$. For each pixel $i\in\mathcal{E}$ on the boundary, we extract a pair of asymmetric samples along its normal vector: one closer to the boundary ($i^+$, the easily confused point) and one farther away ($i^-$, the high-confidence point). If the model assigns consistent semantic categories to points on both sides of the boundary, a penalty is applied:

\begin{equation}
\mathcal{L}_{\text{Boundary}}=\frac{1}{|\mathcal{E}|}\sum_{i\in\mathcal{E}}\left(1+w(i)\cdot\frac{\langle z_s(i^+),\, z_s(i^-)\rangle}{\|z_s(i^+)\|_2\,\|z_s(i^-)\|_2}\right),
\end{equation}
the weight

\begin{equation}
w(i)=\exp(-\ell_{CE}(z_s(i),y_B(i))).
\end{equation}

Adaptive adjustment via cross-entropy loss ensures that strong repulsive forces are applied to proximal blurred pixels only when the reference point prediction is highly accurate. If the reference point itself exhibits considerable uncertainty, the penalty strength exponentially decays, thereby preventing the propagation of erroneous gradients.

Ultimately, the overall objective function for ARCL is as follows.

\begin{equation}
\mathcal{L}_{\text{ARCL}}=\lambda_r\,\mathcal{L}_{\text{Region}}+\lambda_b\,\mathcal{L}_{\text{Boundary}}.
\end{equation}

\subsubsection{Overall Loss and Optimization Objective}

Combining sparse supervised loss, UWSD, TER, and ARCL, the final optimization objective is:

\begin{equation}
\mathcal{L}=\mathcal{L}_{sup}+\lambda_c\mathcal{L}_{UWSD}+\lambda_t\mathcal{L}_{TER}+\mathcal{L}_{ARCL},
\end{equation}
the $\mathcal{L}_{sup}$ is calculated solely on $\Omega_L$, while $\lambda_c$, $\lambda_t$, $\lambda_r$, $\lambda_b$ are used to balance the contributions of the UWSD and TER sub-terms with those of the ARCL sub-term. Additionally, all alignment targets and pseudo labels generated by weak augmentation branches halt gradient backpropagation to ensure training stability.

UWSD, TER, and ARCL are jointly optimized within the same training process, simultaneously improving regional consistency, structural stability, and boundary discriminability under sparse supervision conditions.

\subsection{Network Architecture}

The overall network adopts an encoder-decoder architecture, where the encoder shared by both the strong and weak augmentation branches utilizes DINOv2 as its backbone to leverage its exceptional representational capabilities acquired through large-scale data pre-training. To comprehensively cover the structural spectrum ranging from low-level textures to high-level semantics, we extract intermediate features from the $i \in \{2, 5, 8, 11\}$th layers among all 12 feature layers of DINOv2. This sampling method is a classic configuration of the DPT architecture. In the context of this task, shallow-layer features are sensitive to local contrast and accurately capture color shifts caused by varying layer depths and optical interference from the substrate. Deep-layer features, on the other hand, aggregate global information, abstracting the overall morphology of the material wafer and resisting interference from local surface impurities. This systematic sampling provides multi-scale representations to address the complexity of real-world two-dimensional material micrographs.

The output features of the encoder on the $\ell$th layer are

\begin{equation}
f^{\left(\ell\right)}={\rm \operatorname{Enc}}_\theta^{\left(\ell\right)}(x)\in\operatorname{R}^{H_\ell\times W_\ell\times D_\ell}.
\end{equation}

The multi-level features $\left\{ f^{\left(\ell\right)} \right\}_{\ell=1}^{L}$ undergo $1 \times 1$ convolution projection before being fed into a DPT-based decoder for fusion. The decoder performs deep-to-shallow feature alignment and deep fusion through four levels of RefineNet modules, ultimately upscaling the features by a factor of 14 via bilinear interpolation to generate pixel-level logits:

\begin{equation}
z={\rm
\operatorname{decoder}}_\theta({f^{\left(\ell\right)}})\in\operatorname{R}^{H\times W\times K},\ \ p=Softmax(z).
\end{equation}

This yields the weak-augmentation prediction $p_w$ and the strong-augmentation prediction $p_s$, respectively:

\begin{equation}
p_w=\operatorname{F}_\theta(x^w),\ \ {p}_s=\operatorname{F}_\theta(x^s).
\end{equation}

Network training simultaneously leverages four types of supervisory signals. First, it computes sparse labeling supervision loss on the labeled pixel set $\Omega_L$. Second, it applies UWSD on the unlabeled pixel set $\Omega_U$ to ensure weak/strong augmentation predictions align across unlabeled regions. Third, it introduces TER. This module utilizes a tree structure generated from encoder features that provides structured soft references, delivering online structural constraints and semantic guidance for unlabeled areas. Furthermore, ARCL serves as an essential framework component, enhancing boundary-case handling at both representation and decision levels. On one hand, it constructs augmented labels using weak augmentation branch predictions to expand effective training pixels. On the other hand, it performs sampling and constraints for contrastive learning based on strong augmentation branch probabilities, while combining full-mask construction with boundary supervision on strong augmentation branch logits to impose explicit constraints on boundary neighborhoods. This enhances intra-class cohesion, inter-class separation, and boundary stability.

\section{Experiments}\label{Sec:experiments}

\subsection{Construction and Partitioning of Sparse Annotation Datasets for Two-Dimensional Materials}

To comprehensively evaluate the effectiveness and generalization capability of the proposed method in the task of segmenting microscopic images of two-dimensional materials, this study employed two representative independent datasets: the large-scale multi-scenario segmentation dataset proposed by Yan \cite{Yan2025} and the conventional microscopic segmentation dataset proposed by Uslu \cite{Uslu2024}.

The dataset proposed by Yan encompasses two materials — graphene and MoS$_2$ — and covers multiple fields of view, magnifications, and resolutions, along with typical interfering backgrounds. It represents the largest and most diverse dataset in the field of two-dimensional material flake segmentation to date. It comprises 7,454 microscopic images, including 4,942 graphene images and 2,512 MoS$_2$ images, with approximately 30,000 annotated flake regions. The dataset covers common resolutions such as $1360 \times 1024$, $3088 \times 2056$, and $4272 \times 2848$, featuring a wide range of flake scales. The dataset proposed by Uslu was collected by an automated detection platform and comprises 2,299 microscopic images, including 1,787 graphene images and 512 WSe$_2$ images, all uniformly resolved at 1920×1200 pixels.

Building upon the fully annotated labels provided in the original dataset, this paper constructs a two-dimensional material segmentation scribble annotation dataset using the automated scribble generation paradigm from Scribbles for All \cite{Boettcher2024}. Regarding data processing and partitioning strategies, we implement reasonable adaptations tailored to the characteristics of different data sources: For the Yan dataset, we retain the original mutually exclusive partitioning scheme: graphene training (3,953 images) and validation (989 images), MoS$_2$ training (2,007 images) and validation (505 images). The total dataset comprises 5,960 training images and 1,494 validation images, maintaining an approximate training-to-validation ratio of 4:1. For the Uslu dataset, considering the original multi-class classification task, we uniformly map it to the binary classification task defined in this paper. All materials across all layers are labeled as foreground, while the remainder are labeled as background. Additionally, recognizing the requirement for large training datasets in deep learning models, we swap the small training set and large validation set originally designed for traditional machine learning. The final division is: graphene training set (1,362 images), validation set (425 images); WSe$_2$ training set (420 images), validation set (92 images). The total dataset comprises 1,762 training images and 517 validation images, with a training-to-validation ratio of approximately 7:2.

\begin{figure}[!htbp]
\centering
\includegraphics[width=0.4\linewidth]{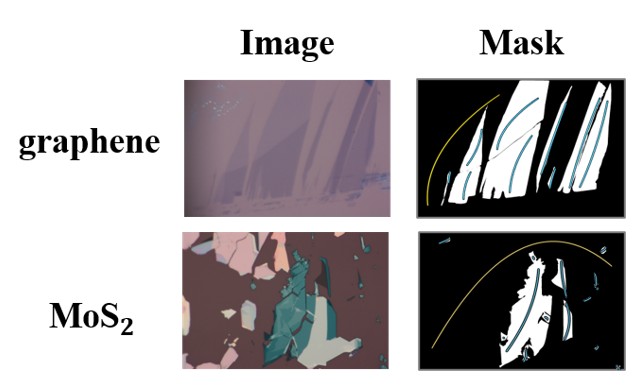}
\caption{\textbf{Illustrative examples of the sparse structural supervision used in our framework.} For each row, the left column shows the original microscopic image. To intuitively display the spatial distribution of the sparse supervision, the scribbles are directly superimposed onto the full masks in the right column. These scribbles capture key geometric structures while avoiding dense annotation, enabling effective learning under extremely limited labeling cost.}
\label{fig:sparse_structure_examples}
\end{figure}

To visually demonstrate the extreme sparsity of scribbles, details the pixel coverage statistics across both training datasets. The data distribution and annotated pixel coverage are shown in \autoref{tab:scribble_coverage}. \autoref{fig:sparse_structure_examples} illustrates examples of scribbles within the dataset. It can be observed that the total pixel coverage of scribbles consistently falls below 0.4\%, fully aligning with the sparse supervision framework characterized by extremely low annotation costs proposed in this paper.

\begin{table*}[!htbp]
  \centering
  \caption{\textbf{Scribble annotation coverage statistics on the graphene and MoS$_2$ training sets.} The metrics report the percentage of labeled pixels relative to the foreground area, background area, and the entire image (total coverage), illustrating the highly sparse nature of the scribble labels used in our experiments. Here, FG denotes foreground pixels, and BG denotes background pixels.}
  \label{tab:scribble_coverage}
  \resizebox{\textwidth}{!}{%
  \begin{tabular}{ccccccc}
    \toprule
    Author & Dataset &  Training Set Size& Foreground Coverage (\%) & Background Coverage (\%) &
    FG/BG Ratio (\%) & Total Coverage (\%) \\
    \midrule
    Yan \cite{Yan2025} & graphene & 3953 & 4.1640 & 0.3053 & 1.4918 & 0.3620 \\
    Yan \cite{Yan2025} & MoS$_2$  & 2007 & 5.0627 & 0.2814 & 1.1952 & 0.3379 \\
    \midrule
    Uslu \cite{Uslu2024} & graphene & 1362 & 23.0045 & 0.5271 & 0.1565 & 0.5622 \\
    Uslu \cite{Uslu2024} & WSe$_2$  & 420 & 30.2252 & 0.5320 & 0.0765 & 0.5547 \\
    \bottomrule
  \end{tabular}%
  }%
\end{table*}

\subsection{Implementation Details}

Our network uses DINOv2 Small as its backbone, training without freezing the backbone parameters. The task is set as binary semantic segmentation, ignoring labels with value 255, and OHEM \cite{Shrivastava2016} is employed to mitigate class imbalance. For data processing, we followed the setup described in the UniMatch V2 framework \cite{Yang2024}. Training inputs are randomly cropped to 518×518 undergo geometric augmentation via random scaling between 0.8 and 1.2, along with random horizontal flipping applied with a 0.5 probability. Weak and strong augmentation views were constructed for consistency learning. Strong augmentation branch additionally applies color and blur jitter with a probability of 0.8. The ColorJitter parameters (brightness, contrast, saturation, hue) are set to 0.2, 0.2, 0.2, and 0.1 respectively; random grayscale probability of 0.05. Gaussian blur probability of 0.2, with the blur radius $\sigma_{\text{blur}}$ randomly sampled within the range of 0.1 to 2.0. These specific data augmentation strategies and hyperparameter ranges are designed to accurately simulate common imaging degradation phenomena in the microscopic characterization of two-dimensional materials, such as color temperature drift caused by aging of the base light source and slight defocusing caused by the movement of the motorized stage. 

In the TEL module, the Gaussian kernel bandwidth parameter $\sigma_{\text{aff}}$ used to construct the minimum spanning tree affinity matrix is set to the empirical value of 0.1. In the ARCL module, the weak confidence threshold $\tau_w$ for pseudo-label filtering is set to a fixed value of 0.7; the strong confidence threshold $\tau_s$ for difficult sample mining is set to a fixed value of 0.97. The temperature coefficient $T$ of the region comparison loss is set to 0.5, with 256 queries and 512 negative keys sampled online for each class.

The optimizer employs AdamW\cite{Loshchilov2018} with a base learning rate of $5 \times 10^{-6}$. Non-backbone parameter groups use a 40x learning rate multiplier. The learning rate follows a poly decay with a fixed power of 0.9. Training runs for 50 epochs with a single-GPU batch size of 2. Two NVIDIA GeForce RTX 3090 24GB GPUs are utilized, with DDP and AMP enabled. The validation phase employed sliding-window inference with a window size of $518 \times 518$ and stride of $448 \times 448$. mIoU served as the evaluation metric, and the model achieving the highest mIoU on the validation set was saved. In addition, the boundary IoU is evaluated simultaneously, and the contour expansion ratio is set to 2\% of the image’s diagonal dimension.

\subsection{Quantitative Comparison with State-of-the-art Methods}

To validate the effectiveness of the proposed method, we first constructed a scribble dataset on the large-scale multi-scenario dataset introduced by Yan \cite{Yan2025} and conducted thorough quantitative evaluations. \autoref{tab:state-of-the-art} presents quantitative results on the graphene and MoS$_2$ datasets, respectively. To ensure fairness in the baseline comparison and the reproducibility of the results, all comparison methods included in the evaluation used the same backbone network, input image dimensions, number of training iterations, optimizer hyperparameters, and validation strategy. As mentioned earlier, existing state-of-the-art scribble-supervised semantic segmentation methods can be categorized into regularized learning (URSS, CC4S), pseudo-label learning (TEL, A$^2$GNN, AGMM, AGMM++), and consistency learning (SASFormer) paradigms. In the context of two-dimensional material microscopic images, to mitigate issues such as noise, under-constraint, and feature confusion arising from sparse annotations, these high-performing representative works often rely on multi-stage training strategies. However, this implicit trade-off comes at the cost of higher training overhead and engineering complexity. In contrast, the proposed TACoS employs a single-stage end-to-end training approach that not only avoids complex workflows but also demonstrates superior performance over multi-stage baseline methods.

\begin{table*}[!htbp]
  \centering
  \caption{\textbf{Quantitative evaluation against state-of-the-art weakly supervised methods on the graphene and MoS$_2$ datasets.} \textbf{Bold} and \underline{underlined} text indicate the best and second-best results under scribble supervision, respectively. A $\checkmark$ in the ``Multi-stage'' column denotes multi-stage training paradigms (e.g., iterative pseudo-label propagation), while its absence indicates single-stage end-to-end training. Our single-stage TACoS framework achieves the highest mIoU and boundary IoU, surpassing both single-stage and complex multi-stage baselines.}
  \label{tab:state-of-the-art}
  \resizebox{\textwidth}{!}{%
  \begin{tabular}{@{}cccc c cccc@{}}
  \toprule
    Method & Publication & Backbone & Supervision & Multi-stage & mIoU  (graphene, \%) & mIoU (MoS$_2$,  \%) &  boundary IoU  (graphene, \%) & boundary IoU (MoS$_2$,  \%)\\\midrule
    Baseline\cite{Ranftl2021}     & CoRR'22 & DINOv2            & Full     &  & 86.33  & 88.35 & 79.52 & 76.76\\
    DCNv4\cite{Xiong2024}   & CVPR'24 & FlashInternImage  & Full     &  & 86.07  & 88.04 & 77.93 & 75.16\\\midrule
    Baseline\cite{Ranftl2021}     & CoRR'22 & DINOv2            & Scribble &  & 77.86  & 78.78 & 56.52 & 49.65\\
    URSS\cite{Pan2021}    & ICCV'21 & DINOv2            & Scribble & $\checkmark$ & \underline{82.35}  & 81.59 & 63.86 & 52.96\\
    A$^2$GNN\cite{Zhang2022AAGNN}   & PAMI'21 & DINOv2            & Scribble & $\checkmark$ & 78.76  & 78.74 & 54.53 & 47.55\\
    SASFormer\cite{Su2023} & ICME'23 & DINOv2          & Scribble &  & 80.23  & 81.97 & \underline{65.55} & 52.80\\
    DCNv4\cite{Xiong2024}   & CVPR'24 & FlashInternImage  & Scribble &  & 79.42  & 79.73 & 62.65 & 57.57\\
    CC4S\cite{Pan2024}    & PAMI'24 & DINOv2            & Scribble & $\checkmark$ & 81.49  & \underline{83.29} & 61.53 & \underline{61.38}\\
    AGMM\cite{Wu2023}  & CVPR'23 & DINOv2            & Scribble &  & 77.65  & 80.88 & 52.54 & 49.29\\
    \textbf{TACoS (Ours)} & -- & DINOv2      & Scribble &  & \textbf{84.09}  & \textbf{85.20} & \textbf{66.95} & \textbf{62.43}\\ \bottomrule
  \end{tabular}%
  }%
\end{table*}

Under the graphene scribble supervision setting, the baseline method DPT achieves a mIoU of 77.86$\%$ using the DINOv2 backbone with only supervised cross-entropy. The multi-stage training URSS improves performance to 82.35$\%$. Using the same backbone network, our proposed TACoS achieves an mIoU of 84.09$\%$ under single-stage training without requiring additional data, representing a 6.23$\%$ improvement over the baseline and further surpassing URSS by 1.74$\%$. Under the MoS$_2$ scribble supervision setting, the DPT baseline achieves mIoU of 78.78$\%$. Among existing methods, CC4S reaches 83.29$\%$, while TACoS achieves the best performance at 85.20$\%$, improving by 6.42$\%$ over the baseline and surpassing CC4S by 1.91$\%$. 

TACoS substantially narrowed the gap with the fully supervised upper bound on both datasets, falling only 2.24$\%$ and 3.15$\%$ short of fully supervised DPT for graphene and MoS$_2$, respectively. This fully demonstrates that TACoS effectively breaks the implicit trade-off between boundary accuracy and regional consistency in existing methods, maintaining robust segmentation performance and outstanding generalization capabilities even under the stringent constraints of extremely low annotation costs and single-stage training.

To further evaluate the accuracy of edge detection, this paper introduces Boundary IoU \cite{cheng2021} as a supplementary metric. As shown in \autoref{tab:state-of-the-art}, TACoS achieved the best Boundary IoU performance on graphene and MoS$_2$ at 66.95\% and 62.43\%, respectively, outperforming methods such as SASFormer and CC4S, thereby validating the advantages of the ARCL and TER modules in enhancing edge discrimination.

\subsection{Performance evaluation on the dataset proposed by Uslu}

We maintained the training settings and constructed scribble annotations on the two-dimensional material microscopic image dataset comprising graphene and WSe$_2$ proposed by Uslu \cite{Uslu2024}, followed by evaluation. We report three categories of results: DPT baseline as the lower bound, TACoS as the proposed strategy, and fully annotated training as the upper bound reference. \autoref{tab:cross_dataset_uslu} summarizes the results. Compared to the baseline, our method achieved stable gains on both datasets (2.99\% improvement on graphene and 4.66\% improvement on WSe$_2$), and notably narrows the gap with the fully labeled training upper bound. It should be noted that due to the particularly severe imbalance in the number of pixels between foreground and background regions in this dataset, the model's overall performance on the Uslu dataset is notably inferior to its performance on the Yan dataset. Nevertheless, our method still demonstrates a certain degree of effectiveness in addressing such extremely imbalanced data.

\begin{table}[!htbp]
  \centering
  \caption{\textbf{Evaluation on the Uslu dataset.} Our single-stage TACoS framework achieves stable performance gains over the baseline on both graphene and WSe$_2$ target domains, demonstrating superior transferability in weakly supervised settings.}
  \label{tab:cross_dataset_uslu}
  \small
  \setlength{\tabcolsep}{6pt}
  \begin{tabular}{l c c c}
    \toprule
     & Baseline & TACoS & Full-Supervision \\
    \midrule
    mIoU (graphene, \%)  & 70.24 & 73.23 & 88.11 \\
    mIoU (WSe$_2$, \%) & 68.41 & 73.07 & 82.41 \\
    boundary IoU (graphene, \%)  & 63.74 & 67.88 & 82.24 \\
    boundary IoU (WSe$_2$, \%) & 61.50 & 68.12 & 79.30 \\
    \bottomrule
  \end{tabular}
\end{table}

Furthermore, on the Boundary IoU metric, TACoS achieved 67.88\% and 68.12\% on graphene and WSe$_2$, respectively — a clear improvement over the baseline (63.74\% and 61.50\%) — demonstrating that the model retains excellent physical boundary preservation capabilities even under extreme class imbalance and complex background interference.

\subsection{Qualitative Demonstration of TACoS and State-of-the-art Methods}

\begin{figure*}[!htbp]
    \centering
    \includegraphics[width=0.6\textwidth]{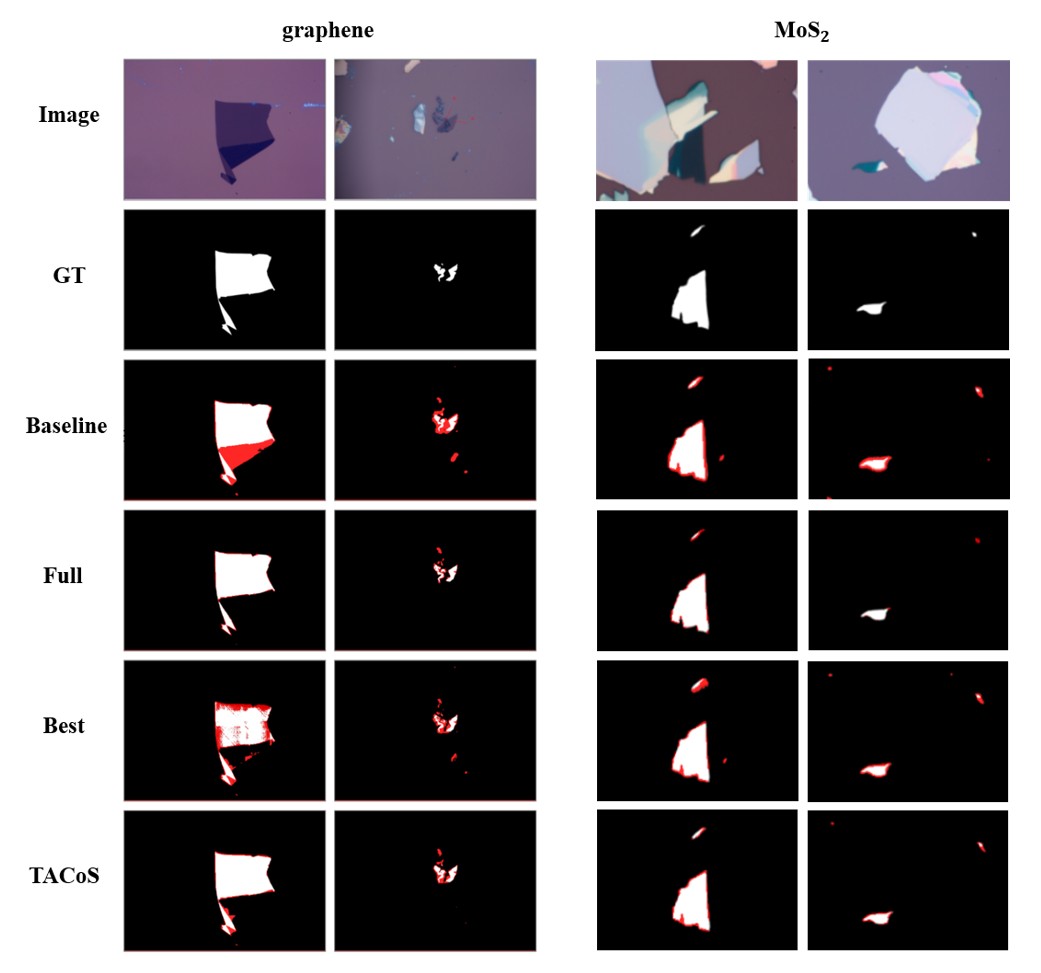}
    \caption{\textbf{Qualitative results of flake segmentation on the Yan dataset.} Visual comparison of segmentation performance for graphene and MoS$_2$ flakes. From top to bottom, the rows display the input optical microscopic images, ground truth (GT) masks, and prediction results from the baseline model, the fully supervised model (Full), comparative methods (URSS for graphene and CC4S for MoS$_2$), and our proposed TACoS framework. Misclassified pixels, including false positives and false negatives, are overlaid in red on the prediction masks. The results demonstrate that TACoS produces cleaner boundaries and substantially fewer segmentation errors compared to the competing methods.}
    \label{fig:qual_graphene_mos2}
\end{figure*}

\begin{figure*}[!htbp]
    \centering
    \includegraphics[width=0.6\textwidth]{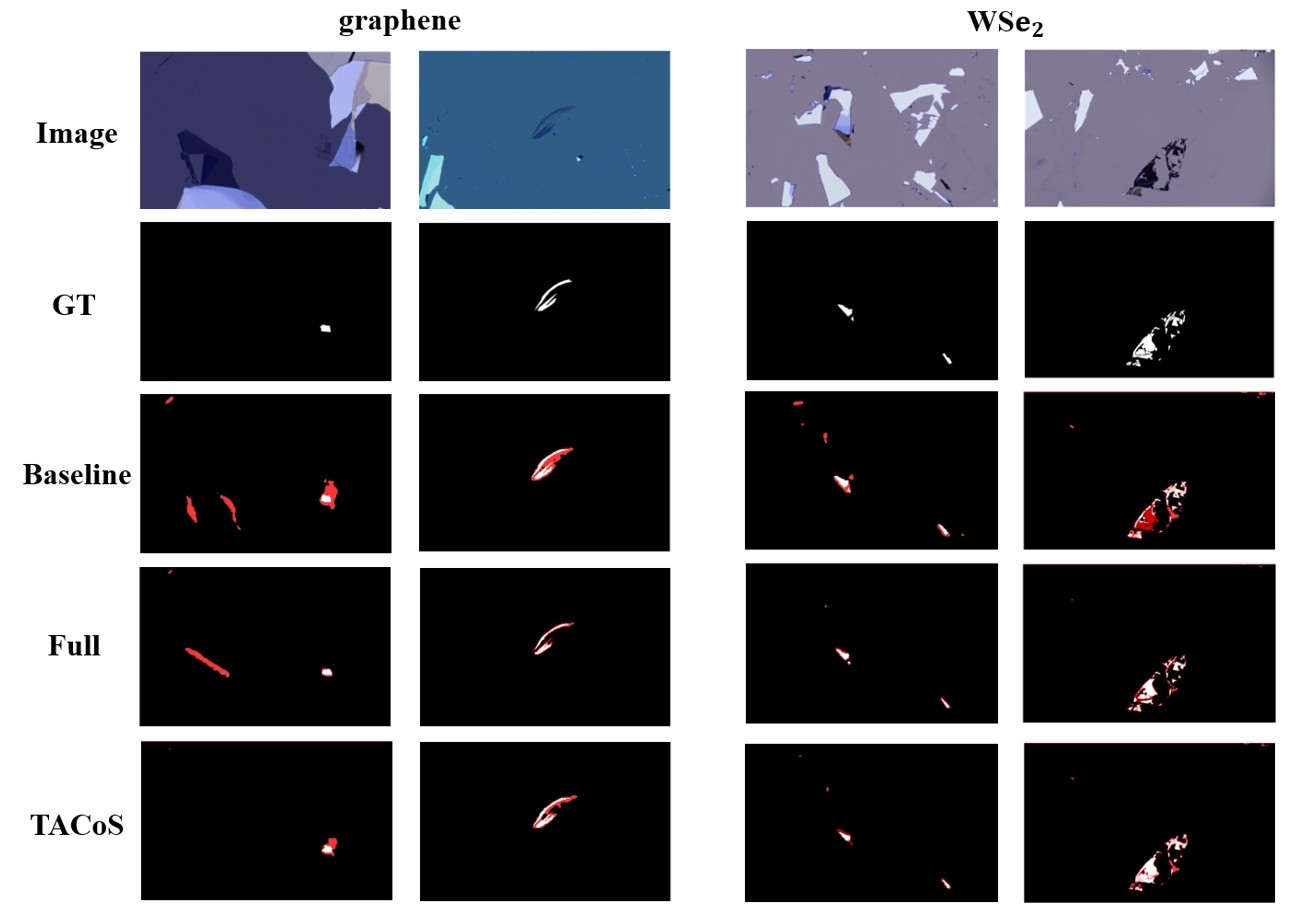}
    \caption{\textbf{Qualitative results of flake segmentation on the Uslu dataset.} Visual comparison of segmentation results on graphene and WSe$_2$ flakes from the Uslu dataset. The rows represent the input microscopic images, ground truth (GT) annotations, and predictions from the baseline model, the full-supervision model (Full), and our TACoS method. Misclassified regions are highlighted in red. TACoS consistently outputs robust segmentation masks with reduced error rates under sparse supervision.}
    \label{fig:qual_graphene_wse2}
\end{figure*}

To visually demonstrate the segmentation performance of the TACoS framework, \autoref{fig:qual_graphene_mos2} and \autoref{fig:qual_graphene_wse2} present qualitative comparisons between our method and existing state-of-the-art approaches on datasets proposed by Yan and Uslu, respectively. Whether on the large-scale multi-scenario dataset containing graphene and MoS$_2$ (\autoref{fig:qual_graphene_mos2}) or on conventional microscopic image datasets (\autoref{fig:qual_graphene_wse2}), TACoS demonstrates substantially superior segmentation performance compared to the baseline model and existing state-of-the-art weakly supervised methods. Visual comparisons reveal that TACoS substantially reduces prediction errors, generating segmentation masks with clearer boundaries and high consistency with ground truth labels. This demonstrates that even under extremely sparse scribble supervision, the method exhibits robust generalization and recognition capabilities, with overall segmentation quality approaching the upper limit of fully supervised benchmarks.

\begin{figure*}[!htbp]
    \centering
    \includegraphics[width=\textwidth]{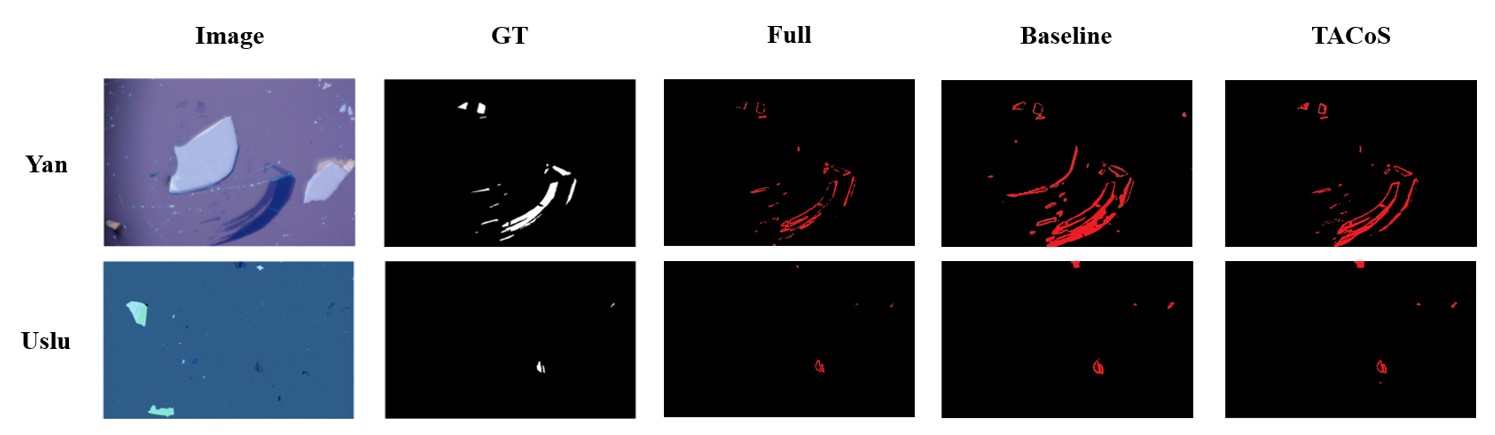}
    \caption{\textbf{Visualization of segmentation failure cases under extreme optical conditions.} The red pixels in the rightmost three columns highlight the misclassified regions (including both false positives and false negatives) produced by the fully supervised model (Full), the scribble-based baseline (Baseline), and the proposed TACoS method, respectively, when compared to the ground truth (GT). The examples are sampled from the Yan and Uslu datasets, illustrating typical boundary shifts and impurity misclassifications induced by weak thin-film interference contrast.}
    \label{fig:fail}
\end{figure*}

Although TACoS substantially bridges the performance gap between weakly supervised and fully supervised models, extreme microscopic imaging conditions still present specific challenges. As shown in \autoref{fig:fail}, misclassifications primarily occur in regions with complex background interference and low optical contrast. The visibility of two-dimensional materials is governed by thin-film interference; when the path length difference is small, the interference contrast drops sharply, blurring the physical boundary between the material edges and the substrate. In sparse labeling, the absence of dense pixel-level boundary constraints reduces the network’s sensitivity to gradient changes along these edges, leading to boundary misalignment. Furthermore, microscopic impurities, substrate defects, or dielectric layers of varying thickness — whose optical properties resemble those of the target material — often trigger false-positive predictions.

\subsection{Ablation Studies}

Given that the dataset proposed by Yan is the most representative in terms of data scale, field-of-view diversity, and complex background interference, all ablation experiments and hyperparameter analyses in this paper are conducted on this dataset to comprehensively validate the effectiveness and operational mechanisms of each core component within the TACoS framework.

\textbf{Loss construction and effectiveness of key terms. }As shown in \autoref{tab:ablation_components}, when using only the sparse supervision baseline, the mIoU for graphene and MoS$_2$ were 77.86\% and 78.78\%, respectively. After introducing UWSD, performance improved to 81.42\% and 81.83\%, corresponding to gains of +3.56\% and +3.05\%. This demonstrates that weak–strong consistency effectively constrains prediction stability in unlabeled regions, constituting the primary source of overall improvement. When only TER is introduced, graphene shows a modest improvement (+0.47\%), while MoS$_2$ demonstrates a more pronounced gain (+3.81\%), indicating that tree energy constraints are more effective at propagating structure to unlabeled pixels in the MoS$_2$ domain. Furthermore, simultaneously enabling UWSD and TER achieves 82.40\% and 82.93\%, respectively, yielding additional gains of +0.98\% and +1.10\% relative to UWSD alone, demonstrating their complementary nature. Introducing ARCL alone yields 81.04\% and 80.19\% (+3.18\% and +1.41\%), confirming that regional-level representation constraints enhance discriminative power. Finally, simultaneously enabling all three achieved optimal scores of 84.09\% and 85.20\%, representing further improvements of +1.69\% and +2.27\% compared to UWSD and TER alone. MoS$_2$ showed more pronounced gains, indicating that ARCL provides stronger representation compensation in more complex and confusing domains.

\begin{table}[!htbp]
  \centering
  \caption{\textbf{Quantitative ablation study of different architectural components within the TACoS framework.} The $\checkmark$ indicates the inclusion of the respective module (UWSD, TER, or ARCL). The results clearly show that the full combination of all proposed modules achieves the optimal mIoU on both the graphene and MoS$_2$ datasets, verifying their mutual effectiveness.}
  \label{tab:ablation_components}
  \small
  \setlength{\tabcolsep}{3pt}
  \begin{tabular}{c c c cc}
    \toprule
    UWSD & TER & ARCL & mIoU  (graphene, \%)& mIoU (MoS$_2$, \%) \\
    \midrule
     &  &  & 77.86 & 78.78 \\
    $\checkmark$ &  &  & 81.42 & 81.83 \\
     & $\checkmark$ &  & 78.33 & 82.59 \\
    $\checkmark$ & $\checkmark$ &  & 82.40 & 82.93 \\
     &  & $\checkmark$ & 81.04 & 80.19 \\
    $\checkmark$ & $\checkmark$ & $\checkmark$ & \textbf{84.09} & \textbf{85.20} \\
    \bottomrule
  \end{tabular}
\end{table}

\textbf{Comparison with Naive EMA and UniMatch v2 Strategies. }Beyond the three components above, we further examine whether teacher-student style strategies yield additional gains. As shown in \autoref{tab:uwsd_variants}, adding either the naive EMA or UniMatch v2 strategy to the UWSD baseline yields slight improvements on graphene (+0.34\% and +0.42\%, respectively), but performance declines on MoS$_2$ (\text{-}0.51\% and \text{-}1.00\%, respectively). This phenomenon indicates that under the material domain and annotation format of this task, directly applying generic EMA or UniMatch v2 perturbation strategies is unstable. Such approaches may amplify pseudo-label noise or cause teacher lag, thereby offsetting potential gains. Therefore, this paper ultimately adopts a task-matched combination of UWSD, TER, and ARCL, rather than relying on naive EMA or directly transferred UniMatch v2 strategies.

\begin{table}[!htbp]
  \centering
  \caption{\textbf{Quantitative comparison of update strategies within the UWSD framework.} The table evaluates the base UWSD against a naive EMA and a UniMatch V2-style method. Results indicate that while advanced EMA techniques offer slight gains on graphene, they fail to generalize effectively to the MoS$_2$ dataset, thereby validating the adoption of our base UWSD architecture for robust single-stage training.}
  \label{tab:uwsd_variants}
  \small
  \setlength{\tabcolsep}{3pt}
  \begin{tabular}{cc c c}
    \toprule
     & UWSD & Naive EMA & UniMatch V2-style \\
    \midrule
    mIoU  (graphene, \%)& 81.42 & 81.76 & 81.84 \\
    mIoU (MoS$_2$, \%)  & 81.83 & 81.32 & 80.83 \\
    \bottomrule
  \end{tabular}
\end{table}

\begin{figure*}[!htbp]
\centering

\includegraphics[width=0.3\textwidth]{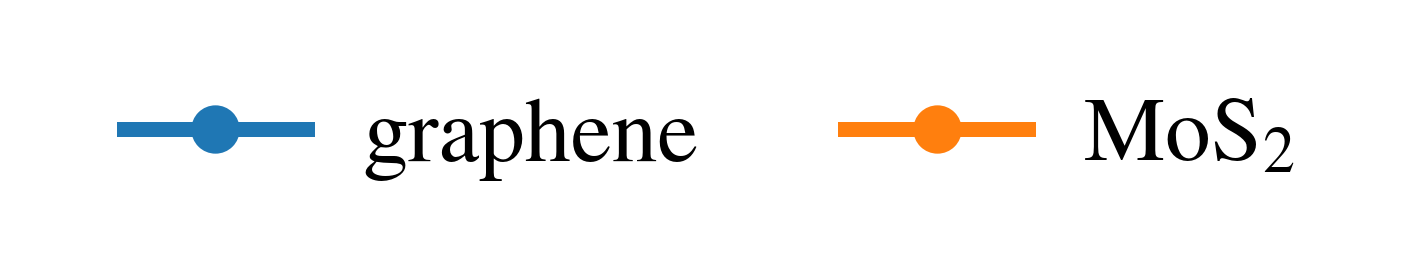}

\subfloat[UWSD $\lambda_c$\label{fig:hparam_lc}]{
\includegraphics[width=0.24\textwidth]{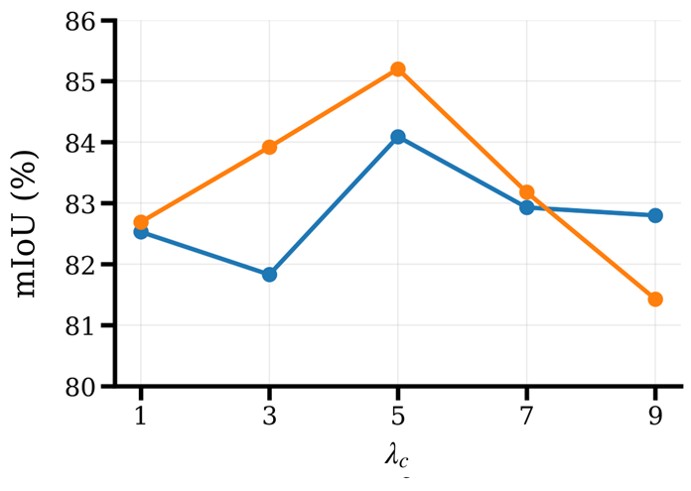}}
\hfill
\subfloat[TER $\lambda_t$\label{fig:hparam_lt}]{
\includegraphics[width=0.24\textwidth]{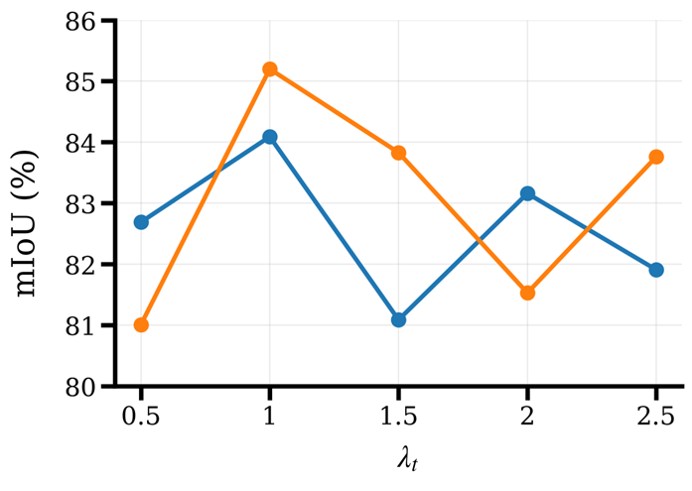}}
\hfill
\subfloat[Region $\lambda_r$\label{fig:hparam_lr}]{
\includegraphics[width=0.24\textwidth]{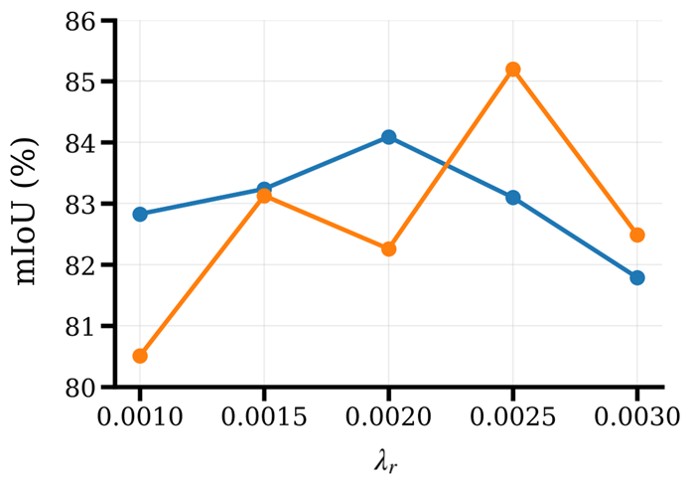}}
\hfill
\subfloat[Boundary $\lambda_b$\label{fig:hparam_lb}]{
\includegraphics[width=0.24\textwidth]{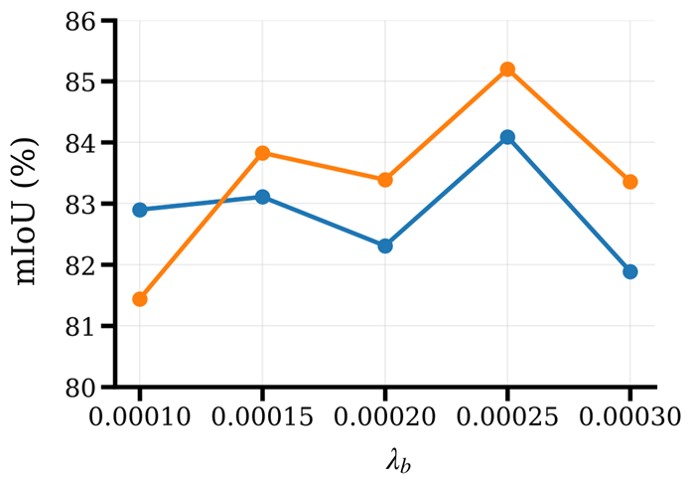}}

\caption{\textbf{Sensitivity analysis of the key hyper-parameters in TACoS.}
The line charts illustrate the fluctuations in mIoU on the graphene and MoS$_2$ datasets with respect to varying values of (a) $\lambda_c$, (b) $\lambda_t$, (c) $\lambda_r$, and (d) $\lambda_b$.}
\label{fig:hparam_sensitivity}

\end{figure*}

\textbf{Hyperparameter analysis. }We performed grid searches on the weights of each loss term across multiple candidate values, reporting representative combinations and results in \autoref{fig:hparam_sensitivity}. To avoid evaluation bias, all hyperparameters were selected solely on the training and validation sets: the fixed network architecture, backbone network, data augmentation, and training budget remained constant. The optimal combination selected on the validation set based on the primary metric mIoU was adopted as the final configuration. Considering the substantial differences in contrast distribution, boundary discernibility, and background texture interference between graphene and MoS$_2$ in optical microscopic images, we report the weight searches and optimal settings for each material separately to reflect the method's stable operating range across different material domains. Overall, the final adopted hyperparameters are:

\begin{itemize}
\item graphene: $\lambda_c=5,\ \lambda_t=1,\ \lambda_r=0.002,\ \lambda_b=0.00025$.

\item MoS$_2$: $\lambda_c=5,\ \lambda_t=1,\ \lambda_r=0.0025,\ \lambda_b=0.00025$.
\end{itemize}

The difference of several orders of magnitude between these loss weights acts as an inversely proportional scaling mechanism. Specifically, the global loss is computed over a massive unlabeled grid, providing comparatively diffuse pixel-level updating signals; conversely, the local loss focuses penalization specifically on sparse, hard-to-classify regions and edges. Assigning a smaller weight to the local loss prevents these concentrated local updates from overshadowing the smooth global alignment, thereby ensuring balanced optimization.

\section{Conclusion}\label{Sec:conclusion}

This paper presents TACoS, a novel, single-stage framework designed to address the substantial bottleneck of expensive pixel-level annotation in the automated high-throughput screening of two-dimensional materials. By synergistically integrating weak-strong consistency learning, tree-energy regularization, and asymmetric regional contrastive learning, TACoS effectively propagates structural and semantic information from highly sparse scribble annotations to unlabelled regions.

The primary strength of our approach lies in its remarkable annotation efficiency and boundary precision. TACoS achieves over 96\% of the performance of fully supervised methods using less than 0.6\% annotated data. For the materials science community, this provides a highly scalable and practical solution. Researchers can now deploy automated flake identification systems without the prohibitive cost of dense manual labeling, thus substantially accelerating the discovery and characterization of novel two-dimensional materials and heterostructures.

Despite these advantages, our framework has limitations. In extreme scenarios characterized by exceedingly low optical contrast or highly complex, textured substrates, the model can still produce false positives or false negatives. The asymmetric regional contrastive learning (ARCL) module relies on relatively clear morphological boundaries; thus, when background impurities perfectly mimic the contrast profiles of target flakes, localized feature confusion can still occur.

Moving forward, future research will focus on extending the TACoS framework to handle multi-class instance segmentation, particularly for identifying overlapping heterostructures of different two-dimensional materials. Additionally, exploring active learning strategies to intelligently suggest the most informative scribble locations could further minimize annotation costs. Finally, extending TACoS to advanced imaging modalities such as Scanning Electron Microscope (SEM) or Atomic Force Microscope (AFM) represents a highly promising path toward a universal foundational model for material segmentation. To integrate these new domains, we propose shifting the distance metric in SEM images from optical color to high-frequency texture descriptors, and directly incorporating physical height differences in AFM data to construct three-dimensional topological constraints that are strictly aligned with actual atomic steps.

\section*{Author Contributions}
\textbf{Jiabei Chen:} Conceptualization, Methodology, Software, Formal analysis, Investigation, Data curation, Visualization, Writing – original draft.
\textbf{Liping Zhang:} Conceptualization, Methodology, Supervision, Writing – review \& editing.
\textbf{Jiang-Bin Wu:} Resources, Data curation.
\textbf{Zhongming Wei:} Resources, Data curation.
\textbf{Enhao Ning:} Writing – review \& editing.
\textbf{Su Yan:} Methodology, Software.
\textbf{Weijun Li:} Supervision, Project administration.
\textbf{Ping-Heng Tan:} Resources, Data curation.
\textbf{Xin Ning:} Conceptualization, Supervision, Project administration, Funding acquisition, Writing – review \& editing.

\section*{Declaration of competing interest}
The authors declare that they have no known competing financial interests or personal relationships that could have appeared to influence the work reported in this paper.

\section*{Acknowledgments}
This work was supported by the National Key Research and Development Program of China (Grant No. 2024YFA1409700) and the CAS Project for Young Scientists in Basic Research (Grant No. YSBR-090).

\section*{Data and Code Availability}
The dataset and code generated in this study will be made publicly available upon the acceptance of this manuscript. The repository will include the full code architecture, PyTorch Lightning implementation, YAML configuration files, and necessary training protocols to ensure the absolute reproducibility of the reported results. Access links will be updated in the final version of the manuscript.


\bibliographystyle{elsarticle-num}  
\bibliography{ref}                 

@article{Geim2007,
  title = {The rise of graphene},
  volume = {6},
  ISSN = {1476-4660},
  number = {3},
  journal = {Nature Materials},
  publisher = {Springer Science and Business Media LLC},
  author = {Geim, A. K. and Novoselov, K. S.},
  year = {2007},
  month = mar,
  pages = {183--191}
}

@article{Novoselov2004,
  title = {Electric Field Effect in Atomically Thin Carbon Films},
  volume = {306},
  ISSN = {1095-9203},
  number = {5696},
  journal = {Science},
  publisher = {American Association for the Advancement of Science (AAAS)},
  author = {Novoselov, K. S. and Geim, A. K. and Morozov, S. V. and Jiang, D. and Zhang, Y. and Dubonos, S. V. and Grigorieva, I. V. and Firsov, A. A.},
  year = {2004},
  month = oct,
  pages = {666--669}
}

@article{Blake2007,
  title = {Making graphene visible},
  volume = {91},
  ISSN = {1077-3118},
  number = {6},
  journal = {Applied Physics Letters},
  publisher = {AIP Publishing},
  author = {Blake, P. and Hill, E. W. and Castro Neto, A. H. and Novoselov, K. S. and Jiang, D. and Yang, R. and Booth, T. J. and Geim, A. K.},
  year = {2007},
  month = aug 
}

@article{Dong2022,
  title = {Deep-Learning-Based Microscopic Imagery Classification, Segmentation, and Detection for the Identification of {2D} Semiconductors},
  volume = {5},
  ISSN = {2513-0390},
  number = {9},
  journal = {Advanced Theory and Simulations},
  publisher = {Wiley},
  author = {Dong, Xingchen and Li, Hongwei and Yan, Yuntian and Cheng, Haoran and Zhang, Hui Xin and Zhang, Yucheng and Le, Tien Dat and Wang, Kun and Dong, Jie and Jakobi, Martin and Yetisen, Ali K. and Koch, Alexander W.},
  year = {2022},
  pages = {2200140},
  month = jul 
}

@article{Sterbentz2021,
  title = {Universal image segmentation for optical identification of {2D} materials},
  volume = {11},
  ISSN = {2045-2322},
  number = {1},
  journal = {Scientific Reports},
  publisher = {Springer Science and Business Media LLC},
  author = {Sterbentz, Randy M. and Haley, Kristine L. and Island, Joshua O.},
  year = {2021},
  pages = {5808},
  month = mar 
}

@inproceedings{Lin2016,
  title = {{ScribbleSup}: Scribble-Supervised Convolutional Networks for Semantic Segmentation},
  booktitle = {Proceedings of the IEEE conference on computer vision and pattern recognition (CVPR)},
  publisher = {IEEE},
  author = {Lin, Di and Dai, Jifeng and Jia, Jiaya and He, Kaiming and Sun, Jian},
  year = {2016},
  month = jun,
  pages = {3159--3167}
}

@inbook{Can2018,
  title = {Learning to Segment Medical Images with Scribble-Supervision Alone},
  ISBN = {9783030008895},
  ISSN = {1611-3349},
  booktitle = {International Workshop on Deep Learning in Medical Image Analysis},
  publisher = {Springer International Publishing},
  author = {Can, Yigit B. and Chaitanya, Krishna and Mustafa, Basil and Koch, Lisa M. and Konukoglu, Ender and Baumgartner, Christian F.},
  year = {2018},
  pages = {236--244}
}

@inproceedings{Pan2021,
  title = {Scribble-Supervised Semantic Segmentation by Uncertainty Reduction on Neural Representation and Self-Supervision on Neural Eigenspace},
  booktitle = {Proceedings of the IEEE/CVF International Conference on Computer Vision (ICCV)},
  publisher = {IEEE},
  author = {Pan, Zhiyi and Jiang, Peng and Wang, Yunhai and Tu, Changhe and Cohn, Anthony G.},
  year = {2021},
  month = oct,
  pages = {7396--7405}
}

@article{Pan2024,
  title = {{CC4S}: Encouraging Certainty and Consistency in Scribble-Supervised Semantic Segmentation},
  volume = {46},
  ISSN = {1939-3539},
  number = {12},
  journal = {IEEE Transactions on Pattern Analysis and Machine Intelligence},
  publisher = {Institute of Electrical and Electronics Engineers (IEEE)},
  author = {Pan, Zhiyi and Sun, Haochen and Jiang, Peng and Li, Ge and Tu, Changhe and Ling, Haibin},
  year = {2024},
  month = dec,
  pages = {8918--8935}
}

@inproceedings{Liang2022,
  title = {{Tree Energy Loss}: Towards Sparsely Annotated Semantic Segmentation},
  booktitle = {Proceedings of the IEEE conference on computer vision and pattern recognition (CVPR)},
  publisher = {IEEE},
  author = {Liang, Zhiyuan and Wang, Tiancai and Zhang, Xiangyu and Sun, Jian and Shen, Jianbing},
  year = {2022},
  month = jun,
  pages = {16886--16895}
}

@article{Zhang2022AAGNN,
  title = {Affinity Attention Graph Neural Network for Weakly Supervised Semantic Segmentation},
  volume = {44},
  ISSN = {1939-3539},
  number = {11},
  journal = {IEEE Transactions on Pattern Analysis and Machine Intelligence},
  publisher = {Institute of Electrical and Electronics Engineers (IEEE)},
  author = {Zhang, Bingfeng and Xiao, Jimin and Jiao, Jianbo and Wei, Yunchao and Zhao, Yao},
  year = {2022},
  month = nov,
  pages = {8082--8096}
}

@inproceedings{Wu2023,
  title = {Sparsely Annotated Semantic Segmentation with Adaptive Gaussian Mixtures},
  booktitle = {Proceedings of the IEEE conference on computer vision and pattern recognition (CVPR)},
  publisher = {IEEE},
  author = {Wu, Linshan and Zhong, Zhun and Fang, Leyuan and He, Xingxin and Liu, Qiang and Ma, Jiayi and Chen, Hao},
  year = {2023},
  month = jun,
  pages = {15454--15464}
}

@article{Wu2025,
  title = {Modeling the Label Distributions for Weakly-Supervised Semantic Segmentation},
  volume = {47},
  ISSN = {1939-3539},
  number = {8},
  journal = {IEEE Transactions on Pattern Analysis and Machine Intelligence},
  publisher = {Institute of Electrical and Electronics Engineers (IEEE)},
  author = {Wu, Linshan and Zhong, Zhun and Ma, Jiayi and Wei, Yunchao and Chen, Hao and Fang, Leyuan and Li, Shutao},
  year = {2025},
  month = aug,
  pages = {6290--6306}
}

@inproceedings{Su2023,
  title = {{SASFormer}: Transformers for Sparsely Annotated Semantic Segmentation},
  booktitle = {Proceedings of the IEEE International Conference on Multimedia and Expo (ICME)},
  publisher = {IEEE},
  author = {Su, Hui and Ye, Yue and Hua, Wei and Cheng, Lechao and Song, Mingli},
  year = {2023},
  month = jul,
  pages = {390--395}
}

@article{Yan2025,
  title = {Large scale and diverse two-dimensional flake segmentation dataset by general-purpose and labor-efficient annotation framework},
  volume = {9},
  ISSN = {2397-7132},
  number = {1},
  journal = {npj 2D Materials and Applications},
  publisher = {Springer Science and Business Media LLC},
  author = {Yan, Su and Chen, Jun and Li, Xiao and Bai, Jinfan and Zhou, Rong and Zhang, Liping and Li, Weijun and Wu, Jiang-Bin and Tan, Ping-Heng and Ning, Xin},
  year = {2025},
  pages = {103},
  month = nov 
}

@article{Oquab2024,
  title = {{DINOv2}: Learning Robust Visual Features without Supervision},
  author = {Maxime Oquab and Timoth{\'e}e Darcet and Th{\'e}o Moutakanni and Huy V. Vo and Marc Szafraniec and Vasil Khalidov and Pierre Fernandez and Daniel HAZIZA and Francisco Massa and Alaaeldin El-Nouby and Mido Assran and Nicolas Ballas and Wojciech Galuba and Russell Howes and Po-Yao Huang and Shang-Wen Li and Ishan Misra and Michael Rabbat and Vasu Sharma and Gabriel Synnaeve and Hu Xu and Herve Jegou and Julien Mairal and Patrick Labatut and Armand Joulin and Piotr Bojanowski},
  journal = {arXiv preprint arXiv:2304.07193},
  year = {2023},
}

@inproceedings{Ranftl2021,
  title = {Vision Transformers for Dense Prediction},
  booktitle = {Proceedings of the IEEE/CVF International Conference on Computer Vision (ICCV)},
  publisher = {IEEE},
  author = {Ranftl, Rene and Bochkovskiy, Alexey and Koltun, Vladlen},
  year = {2021},
  month = oct,
  pages = {12159--12168}
}

@inproceedings{Xiong2024,
  title = {Efficient Deformable {ConvNets}: Rethinking Dynamic and Sparse Operator for Vision Applications},
  booktitle = {Proceedings of the IEEE conference on computer vision and pattern recognition (CVPR)},
  publisher = {IEEE},
  author = {Xiong, Yuwen and Li, Zhiqi and Chen, Yuntao and Wang, Feng and Zhu, Xizhou and Luo, Jiapeng and Wang, Wenhai and Lu, Tong and Li, Hongsheng and Qiao, Yu and Lu, Lewei and Zhou, Jie and Dai, Jifeng},
  year = {2024},
  month = jun,
  pages = {5652--5661}
}

@article{Masubuchi2020,
  title = {Deep-learning-based image segmentation integrated with optical microscopy for automatically searching for two-dimensional materials},
  volume = {4},
  ISSN = {2397-7132},
  number = {1},
  journal = {npj 2D Materials and Applications},
  publisher = {Springer Science and Business Media LLC},
  author = {Masubuchi, Satoru and Watanabe, Eisuke and Seo, Yuta and Okazaki, Shota and Sasagawa, Takao and Watanabe, Kenji and Taniguchi, Takashi and Machida, Tomoki},
  year = {2020},
  pages = {3},
  month = mar 
}

@article{Uslu2024,
  title = {An open-source robust machine learning platform for real-time detection and classification of {2D} material flakes},
  volume = {5},
  ISSN = {2632-2153},
  number = {1},
  journal = {Machine Learning: Science and Technology},
  publisher = {IOP Publishing},
  author = {Uslu, Jan-Lucas and Ouaj, Taoufiq and Tebbe, David and Nekrasov, Alexey and Bertram, Jo Henri and Sch\"{u}tte, Marc and Watanabe, Kenji and Taniguchi, Takashi and Beschoten, Bernd and Waldecker, Lutz and Stampfer, Christoph},
  year = {2024},
  month = feb,
  pages = {015027}
}

@article{Zhang2022Coatings,
  title = {Deep Learning-Based Layer Identification of {2D} Nanomaterials},
  volume = {12},
  ISSN = {2079-6412},
  number = {10},
  journal = {Coatings},
  publisher = {MDPI AG},
  author = {Zhang, Yu and Zhang, Heng and Zhou, Shujuan and Liu, Guangjie and Zhu, Jinlong},
  year = {2022},
  month = oct,
  pages = {1551}
}

@article{Yang2024,
  title = {Identification and Structural Characterization of Twisted Atomically Thin Bilayer Materials by Deep Learning},
  volume = {24},
  ISSN = {1530-6992},
  number = {9},
  journal = {Nano Letters},
  publisher = {American Chemical Society (ACS)},
  author = {Yang, Haitao and Hu, Ruiqi and Wu, Heng and He, Xiaolong and Zhou, Yan and Xue, Yizhe and He, Kexin and Hu, Wenshuai and Chen, Haosen and Gong, Mingming and Zhang, Xin and Tan, Ping-Heng and Hernández, Eduardo R. and Xie, Yong},
  year = {2024},
  month = feb,
  pages = {2789--2797}
}

@inproceedings{He2017,
  title = {{Mask R-CNN}},
  booktitle = {Proceedings of the IEEE/CVF International Conference on Computer Vision (ICCV)},
  publisher = {IEEE},
  author = {He, Kaiming and Gkioxari, Georgia and Dollar, Piotr and Girshick, Ross},
  year = {2017},
  month = oct,
  pages = {2980--2988}
}

@inproceedings{Ouali2020,
  title = {Semi-Supervised Semantic Segmentation With Cross-Consistency Training},
  booktitle = {Proceedings of the IEEE Conference on Computer Vision and Pattern Recognition (CVPR)},
  publisher = {IEEE},
  author = {Ouali, Yassine and Hudelot, Celine and Tami, Myriam},
  year = {2020},
  month = jun,
  pages = {12671--12681}
}

@inproceedings{Chen2021,
  author = {Chen, Xiaokang and Yuan, Yuhui and Zeng, Gang and Wang, Jingdong},
  title = {Semi-Supervised Semantic Segmentation With Cross Pseudo Supervision},
  booktitle = {Proceedings of the IEEE Conference on Computer Vision and Pattern Recognition (CVPR)},
  month = jun,
  year = {2021},
  pages = {2613--2622}
}

@inproceedings{Wang2021,
  title = {Exploring Cross-Image Pixel Contrast for Semantic Segmentation},
  booktitle = {Proceedings of the IEEE/CVF international conference on computer vision (ICCV)},
  publisher = {IEEE},
  author = {Wang, Wenguan and Zhou, Tianfei and Yu, Fisher and Dai, Jifeng and Konukoglu, Ender and Gool, Luc Van},
  year = {2021},
  month = oct,
  pages = {7283--7293}
}

@article{Hamilton2022,
  title = {Unsupervised Semantic Segmentation by Distilling Feature Correspondences},
  author = {Mark Hamilton and Zhoutong Zhang and Bharath Hariharan and Noah Snavely and William T. Freeman},
  journal={arXiv preprint arXiv:2203.08414},
  year = {2022}
}

@inproceedings{Sohn2020,
  title = {{FixMatch}: Simplifying Semi-Supervised Learning with Consistency and Confidence},
  author = {Sohn, Kihyuk and Berthelot, David and Carlini, Nicholas and Zhang, Zizhao and Zhang, Han and Raffel, Colin and Cubuk, Ekin Dogus and Kurakin, Alex and Li, Chun-Liang},
  booktitle = {Advances in Neural Information Processing Systems (NeurIPS)},
  volume = {33},
  pages = {596--608},
  year = {2020}
}

@article{Yang2025,
  title = {{UniMatch V2}: Pushing the Limit of Semi-Supervised Semantic Segmentation},
  volume = {47},
  ISSN = {1939-3539},
  number = {4},
  journal = {IEEE Transactions on Pattern Analysis and Machine Intelligence},
  publisher = {Institute of Electrical and Electronics Engineers (IEEE)},
  author = {Yang, Lihe and Zhao, Zhen and Zhao, Hengshuang},
  year = {2025},
  month = apr,
  pages = {3031--3048}
}

@article{Liu2022,
  title = {Bootstrapping Semantic Segmentation with Regional Contrast},
  author = {Shikun Liu and Shuaifeng Zhi and Edward Johns and Andrew Davison},
  journal={arXiv preprint arXiv:2104.04465},
  year = {2021}
}

@article{Qiu2024,
  title = {Guided contrastive boundary learning for semantic segmentation},
  volume = {155},
  ISSN = {0031-3203},
  journal = {Pattern Recognition},
  publisher = {Elsevier BV},
  author = {Qiu, Shoumeng and Chen, Jie and Zhang, Haiqiang and Wan, Ru and Xue, Xiangyang and Pu, Jian},
  year = {2024},
  month = nov,
  pages = {110723}
}

@inproceedings{Boettcher2024,
  author = {Wolfgang Boettcher and Lukas Hoyer and Ozan Unal and Jan Eric Lenssen and Bernt Schiele},
  title = {{Scribbles for All}: Benchmarking Scribble Supervised Segmentation Across Datasets},
  volume = {37},
  pages = {46002--46024},
  year = {2024},
  booktitle = {Advances in Neural Information Processing Systems (NeurIPS)}
}

@inproceedings{Shrivastava2016,
  author = {Abhinav Shrivastava and Abhinav Gupta and Ross B. Girshick},
  title = {Training Region-Based Object Detectors with Online Hard Example Mining},
  year = {2016},
  pages = {761--769},
  booktitle = {Proceedings of the IEEE Conference on Computer Vision and Pattern Recognition (CVPR)}
}

@article{Loshchilov2018,
  title = {Decoupled Weight Decay Regularization},
  author = {Ilya Loshchilov and Frank Hutter},
  journal = {arXiv preprint arXiv:1711.05101},
  year = {2017}
}

@article{liang2026,
  title={{B$^3$CT}: Three-Branch Learning with Unlabeled Target Signals for Domain-Robust Semantic Segmentation},
  author={Liang, Chen and Zhao, Xin and Jia, Jian and Wang, Junyan and Cao, Lijun and Zhang, Jianguo and Chen, Weihua},
  journal={International Journal of Computer Vision},
  volume={134},
  number={3},
  pages={123},
  year={2026},
  publisher={Springer}
}

@inproceedings{cheng2021,
  title={{Boundary IoU}: Improving object-centric image segmentation evaluation},
  author={Cheng, Bowen and Girshick, Ross and Doll{\'a}r, Piotr and Berg, Alexander C and Kirillov, Alexander},
  booktitle={Proceedings of the IEEE/CVF conference on computer vision and pattern recognition},
  pages={15334--15342},
  year={2021}
}

\end{document}